\documentclass{article}

\PassOptionsToPackage{numbers, compress}{natbib}

\usepackage[preprint]{neurips_2025}

\usepackage[utf8]{inputenc} 
\usepackage[T1]{fontenc}    
\usepackage{hyperref}       
\usepackage{url}            
\usepackage{booktabs}       
\usepackage{amsfonts}       
\usepackage{nicefrac}       
\usepackage{microtype}      
\usepackage[table,usenames,dvipsnames]{xcolor}         

\usepackage{comment}
\usepackage{graphicx}
\usepackage{float}
\title{Quad Length Codes for \\
Lossless Compression of e4m3}

\author{%
  Aditya Agrawal \\
  \texttt{adityaag@google.com} \\
  \And
  Albert Magyar \\
  \texttt{amagyar@google.com} \\
  \And
  Hiteshwar Eswaraiah \\
  \texttt{hiteshwarbe@google.com} \\
  \And
  Patrick Sheridan \\
  \texttt{sheridp@google.com} \\
  \And
  Pradeep Janedula \\
  \texttt{pjanedula@google.com} \\
  \And
  Ravi Krishnan Venkatesan \\
  \texttt{ravikrishnan@google.com} \\
  \And
  Krishna Nair \\
  \texttt{krishnaknair@google.com} \\
  \And
  Ravi Iyer \\
  \texttt{raviyer@google.com} \\
  \AND
  Google LLC
}

\begin{document}
\maketitle
\begin{abstract}
Training and serving Large Language Models (LLMs)
relies heavily on parallelization and collective operations,
which are frequently bottlenecked by network bandwidth.
Lossless compression using e.g., Huffman codes can alleviate
the issue, however, Huffman codes suffer from slow,
bit-sequential decoding and high hardware complexity due
to deep tree traversals.
Universal codes e.g., Exponential-Golomb codes are faster to
decode but do not exploit the symbol frequency distributions.
To address these limitations, this paper introduces
Quad Length Codes, a hybrid approach designed to balance
compression efficiency with decoding speed.
The coding scheme uses 3 prefix bits to
divide the 256 symbols into 8 areas.
Each area has a different code length
and encodes a different number of symbols.
The scheme uses a Look Up Table with
256 entries, significantly simplifying the
hardware implementation compared to Huffman trees.
The coding scheme can be adapted for different 
distributions.
For the e4m3 data type, the scheme achieves
a compressibility of 13.9\%
in comparison to 15.9\% achieved by Huffman codes,
but it significantly speeds up the decoding
and simplifies the hardware complexity.
\end{abstract}


\section{Background}

Training and serving Large Language Models (LLMs) e.g.,
Gemini \cite{models:gemini_2025}, Gemma \cite{models:gemma1,
models:gemma2}, LLaMA \cite{models:llama}, GPT \cite{models:gpt_1}
require partitioning (sharding) the data (parameters, activations,
optimizer state etc.) and parallelizing the computation across
multiple accelerators.
There are multiple paradigms of parallelism e.g., Data Parallelism,
Tensor Parallelism, Pipeline Parallelism, Expert Parallelism and
Sequence Parallelism \cite{parallelism:megatron, parallelism:sequence,
parallelism:zero, parallelism:nvidia_1, parallelism:nvidia_2,
parallelism:colossalai, parallelism:scaling-book}.
Different parallelization strategies invoke different collective
operations e.g., AllReduce, ReduceScatter, AllGather, AlltoAll
\cite{collectives:nccl}. Collective operations are typically
bounded by network bandwidth. Lossless compression is an
effective way to reduce the network traffic and
improve collective performance.

Huffman codes \cite{huffman} either directly or as part of other
algorithms e.g., DEFLATE \cite{deflate}, Zstandard \cite{zstd},
Brotli \cite{brotli} are commonly used for lossless data
compression. They exploit the distribution of symbol frequencies
and are optimal entropy codes.
However, they are variable length and decoding them
requires a binary tree traversal.
The decoding is slow and bit sequential i.e., the decode
latency is proportional to the number of bits in the
encoded input.

Universal codes \cite{universal} e.g., Elias Gamma codes
\cite{gamma-elias}, Elias Delta codes \cite{delta-elias},
Elias Omega codes \cite{omega-elias},
Exponential Golomb codes \cite{exp-golomb} are also
variable length codes.
In these codes, the length of the code is part of the code,
for example, Elias Gamma codes use a series of leading zeros
to indicate the length of the code.
Therefore, the decoding is not completely bit sequential
and hence faster.
However, these codes do not exploit the distribution of symbol
frequencies and hence are not optimal.

\section{Objective}
Our objective is to develop codes such that

\begin{itemize}
\item {They exploit the distribution of symbol frequencies, similar to Huffman codes.}
\item {The decoder is not bit sequential and hence faster, similar to Universal codes.}
\end{itemize}

\section{Experimental Setup}
We analyzed the Gemma 2B model \cite{models:gemma1}
during Supervised Fine Tuning (SFT). The model has
18 layers and is sharded over 64 TPUs.
We analyzed the weight, activation, weight gradient and
activation gradient tensors of the feed forward layers,
FFN1 and FFN2.
Overall, there are 18 x 64 = 1152 shards of each
tensor type e.g., FFN1 activation.
We present our observations and results for
FFN1 and FFN2 activation tensors at
the eXmY \cite{format:exmy} e4m3 data type 
where all 256 encodings are finite.
The OCP e4m3 format \cite{format:ocpmx}
has 2 out of 256 encodings reserved for NaNs,
which will have minimal effect on the symbol
probabilities and the results.
The quantization block size is 32.

\section{Observations}

\begin{figure}[h]
  \centering
  \includegraphics[width=\linewidth]{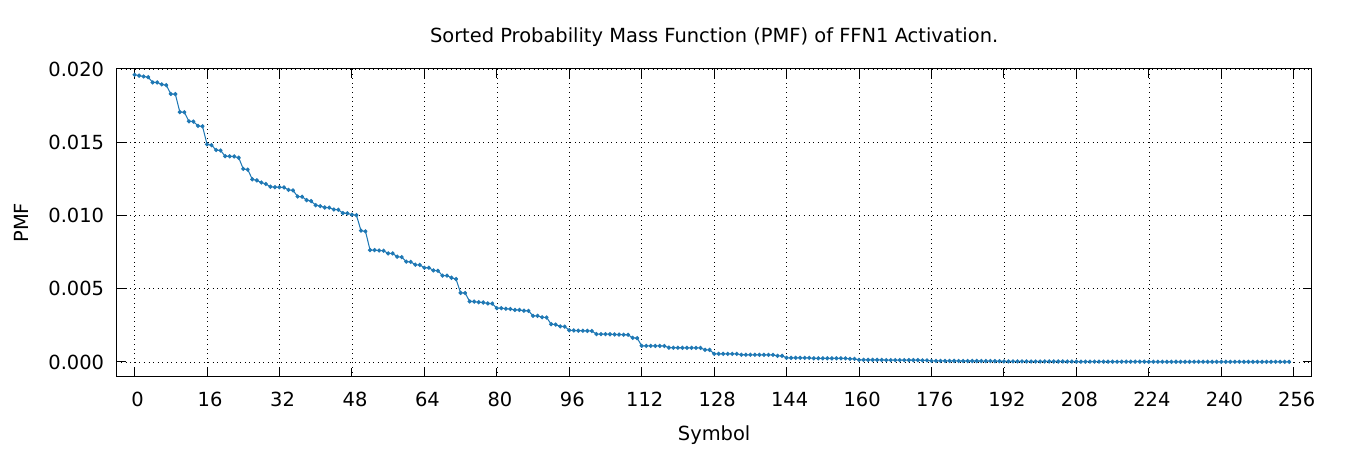}
  \caption{Sorted Probability Mass Function (PMF) of FFN1 activation.}
  \label{fig:ffn1_sorted_pmf}
\end{figure}

Fig. \ref{fig:ffn1_sorted_pmf} shows the Probability
Mass Function (PMF) of  FFN1 activation with a
symbol size of 8 bits i.e., 256 symbols, averaged
over all shards.
The probabilities are sorted in decreasing order.
The histograms of FFN1 weights, FFN1 weight gradient,
FFN2 weights, and FFN2 weight gradient
exhibit similar behaviour.
This distribution has a Shannon entropy \cite{entropy}
of 6.69 bits and hence an ideal compressibility of
$(8-6.69)/8 \approx 16.3\%$.

\begin{figure}[h]
  \centering
  \includegraphics[width=\linewidth]{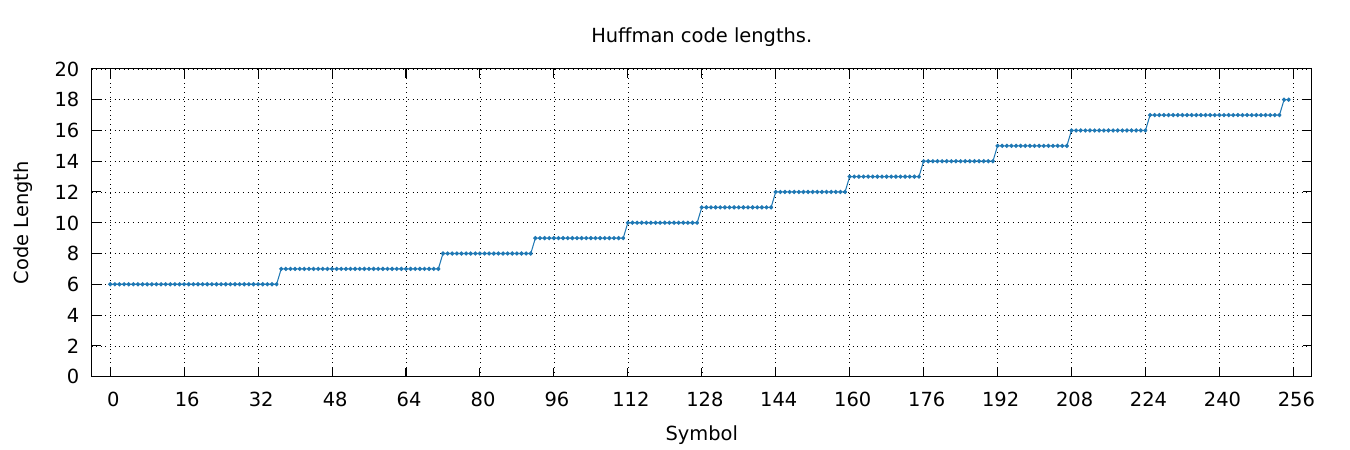}
  \caption{Huffman code lengths for each symbol.}
  \label{fig:ffn1_code_length}
\end{figure}

Fig. \ref{fig:ffn1_code_length} shows the Huffman
\cite{huffman} code lengths of the different
symbols. More frequent symbols have smaller
codes and less frequent symbols have longer codes
as expected. The compressibility achieved using Huffman
codes is 15.9\%. For this distribution, the code
lengths range from 6 to 18. This makes the Huffman tree
very deep and increases the hardware complexity.

\section{Quad Length Codes}

We took a cue from the code lengths assigned
by the Huffman algorithm, as shown in Fig.
\ref{fig:ffn1_code_length}.
We observed that the top 37 symbols have a
code length of 6 bits,
the next 35 symbols have a code length of 7
bits and so on.
The other symbols have code lengths from 8 to 18 bits.
This contributes to the optimal compressibility
achieved by Huffman codes, but also to its complexity.

\begin{figure}[h]
  \centering
  \includegraphics[width=\linewidth]{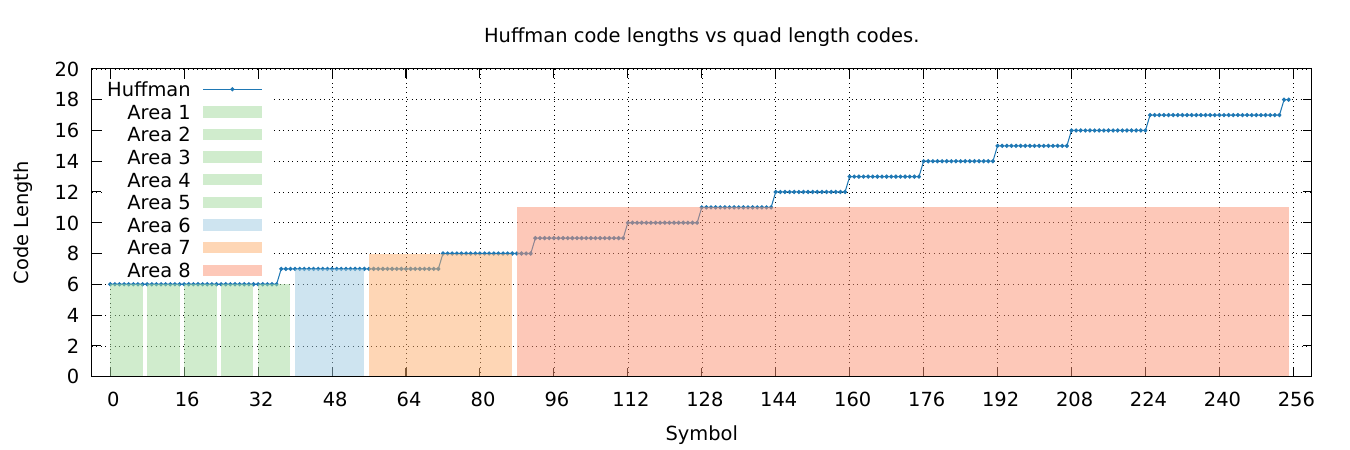}
  \caption{Code lengths for Huffman codes and quad length codes.}
  \label{fig:ffn1_quad_length}
\end{figure}

We divide the 256 symbols into 8 areas. Identifying the area
requires 3 bits. Each area has a different number of symbols.
The first five areas have 8 symbols each and hence require
$3 + 3 = 6$ bits,
the sixth area has 16 symbols and hence requires
$3 + 4 = 7$ bits,
the seventh area has 32 symbols and hence requires
$3 + 5 = 8$ bits,
and the last area has the remaining 168 symbols and hence
requires $3 + 8 = 11$ bits.
This is shown in Fig. \ref{fig:ffn1_quad_length}
and Table \ref{table:ffn1_quad_length}.

Overall, this scheme has codes of 4 different lengths,
hence quad length codes.
The `area codes' i.e., the first three bits, encode the length
of the code which can be either 6, 7, 8, or 11 bits.
In contrast, the Huffman codes shown above have
13 different code lengths.
This simplifies the software or hardware decoder
implementation.
The compressibility achieved using this scheme is 13.9\%
which is about 2.0\% less than Huffman.

\begin{table}[h]
\caption{Quad length coding scheme.}
\renewcommand{\arraystretch}{1.5}
\centering
\rowcolors{1}{RoyalBlue!10}{}
\begin{tabular}{|c|c|c|c|c|c|}
\toprule \hline
\rowcolor{RoyalBlue!40}
Area    & Area code & \#Symbols & \#Symbol bits  & Code length  & Symbol Range \\
1       & 000       & 8         & 3             & 6             & 0-7   \\
2       & 001       & 8         & 3             & 6             & 8-15  \\
3       & 010       & 8         & 3             & 6             & 16-23 \\
4       & 011       & 8         & 3             & 6             & 24-31 \\
5       & 100       & 8         & 3             & 6             & 32-39 \\
6       & 101       & 16        & 4             & 7             & 40-55 \\
7       & 110       & 32        & 5             & 8             & 56-87 \\
8       & 111       & 168       & 8             & 11            & 88-255 \\
\hline \bottomrule
\end{tabular}
\label{table:ffn1_quad_length}
\end{table}

\section{Adaptation}

\begin{figure}[h]
  \centering
  \includegraphics[width=\linewidth]{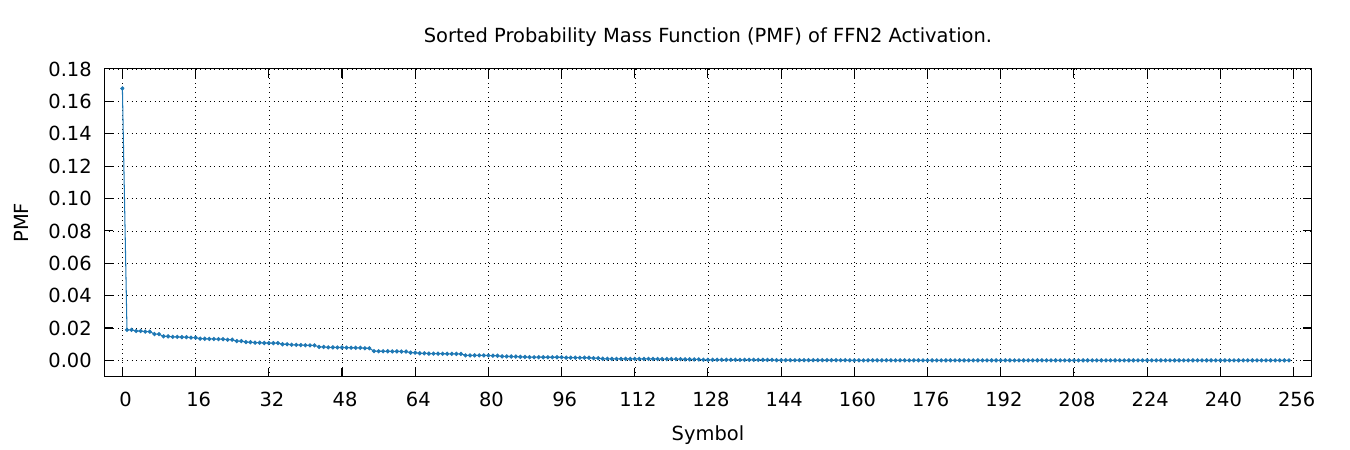}
  \caption{Sorted Probability Mass Function (PMF) of FFN2 activation.}
  \label{fig:ffn2_sorted_pmf}
\end{figure}

Fig. \ref{fig:ffn2_sorted_pmf} shows the PMF of
FFN2 activation. The probabilities are sorted in
decreasing order.
The PMF shows that 1 symbol (zero) occurs with
a significantly higher frequency than other 
symbols. This is due to the intervening
non-linear activation function.
The PMF of FFN1 activation gradient and
FFN2 activation gradient are similar.
The Shannon entropy of this distribution
is 6.11 bits and hence has an ideal 
compressibility of $(8-6.11)/8 = 23.6\%$.

\begin{figure}[h]
  \centering
  \includegraphics[width=\linewidth]{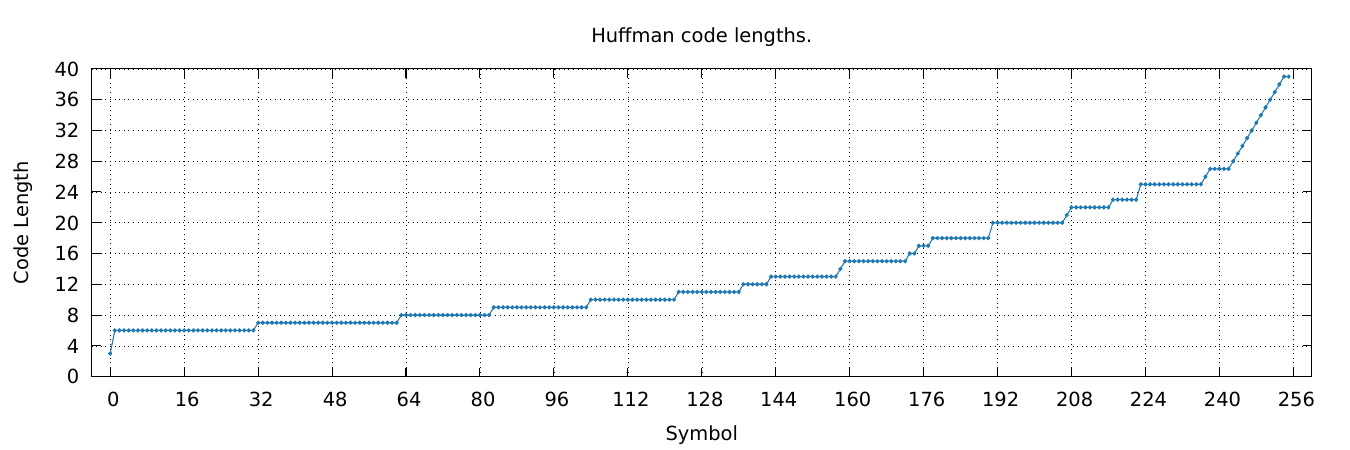}
  \caption{Huffman code lengths for each symbol.}
  \label{fig:ffn2_code_length}
\end{figure}

Fig. \ref{fig:ffn2_code_length} shows the
corresponding Huffman code lengths
of the different symbols.
The code lengths range from 3 to 39 bits.
This makes the Huffman tree very deep
and increases the hardware complexity.
The compressibility achieved using
Huffman codes is $23.2\%$.

Using the quad length coding scheme from
Table \ref{table:ffn1_quad_length} for this
distribution results in a compressibility
of only $16.7\%$.
This is because the frequency distributions
are not similar e.g., for this distribution, the 
most frequent symbol is assigned 
a code length of 3 bits instead of 6 bits,
as in Table \ref{table:ffn1_quad_length}.

\begin{figure}[h]
  \centering
  \includegraphics[width=\linewidth]{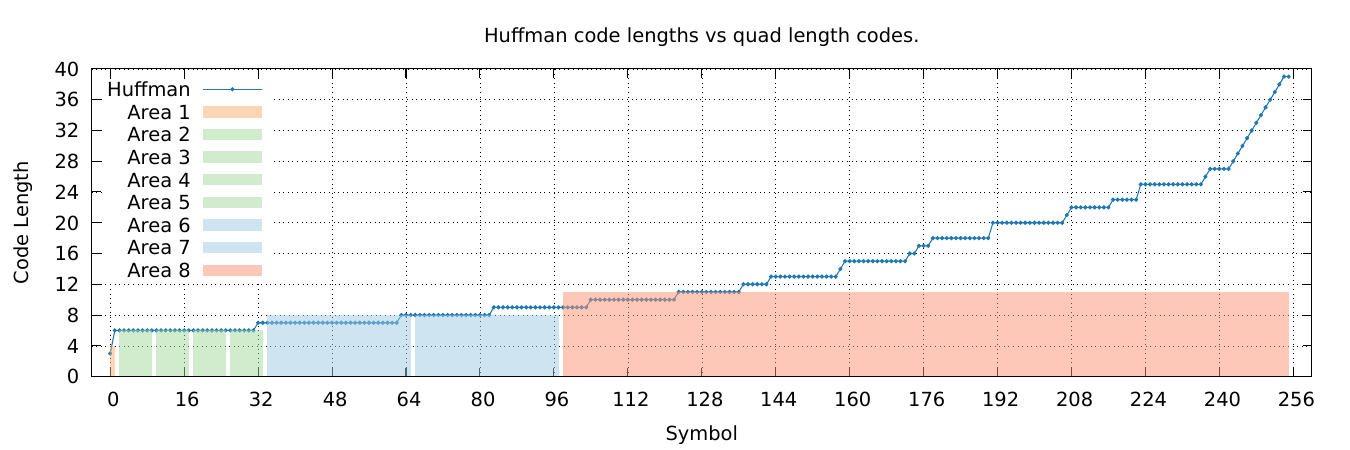}
  \caption{Code lengths for Huffman codes and quad length codes.}
  \label{fig:ffn2_quad_length}
\end{figure}

We can adapt the coding scheme to better 
fit this distribution and is shown in
Table \ref{table:ffn2_quad_length} and
Fig. \ref{fig:ffn2_quad_length}.
This coding scheme also has codes of
4 different lengths viz., 4, 6, 8 and 11 bits.
The compressibility achieved using this
new coding scheme is 19.0\%, an
improvement of $2.3\%$ from the scheme
in Table \ref{table:ffn1_quad_length}
and about $4\%$ less than Huffman.

\begin{table}[h]
\caption{Quad length coding scheme.}
\renewcommand{\arraystretch}{1.5}
\centering
\rowcolors{1}{RoyalBlue!10}{}
\begin{tabular}{|c|c|c|c|c|c|}
\toprule \hline
\rowcolor{RoyalBlue!40}
Area    & Area code & \#Symbols & \#Symbol bits  & Code length  & Symbol Range \\
1       & 000       & 2         & 1             & 4             & 0-1   \\
2       & 001       & 8         & 3             & 6             & 2-9  \\
3       & 010       & 8         & 3             & 6             & 10-17 \\
4       & 011       & 8         & 3             & 6             & 18-25 \\
5       & 100       & 8         & 3             & 6             & 26-33 \\
6       & 101       & 32        & 5             & 8             & 34-65 \\
7       & 110       & 32        & 5             & 8             & 66-97 \\
8       & 111       & 158       & 8             & 11            & 98-255 \\
\hline \bottomrule
\end{tabular}
\label{table:ffn2_quad_length}
\end{table}

\section{Implementation}

\begin{figure}[h]
  \centering
  \includegraphics[width=\linewidth]{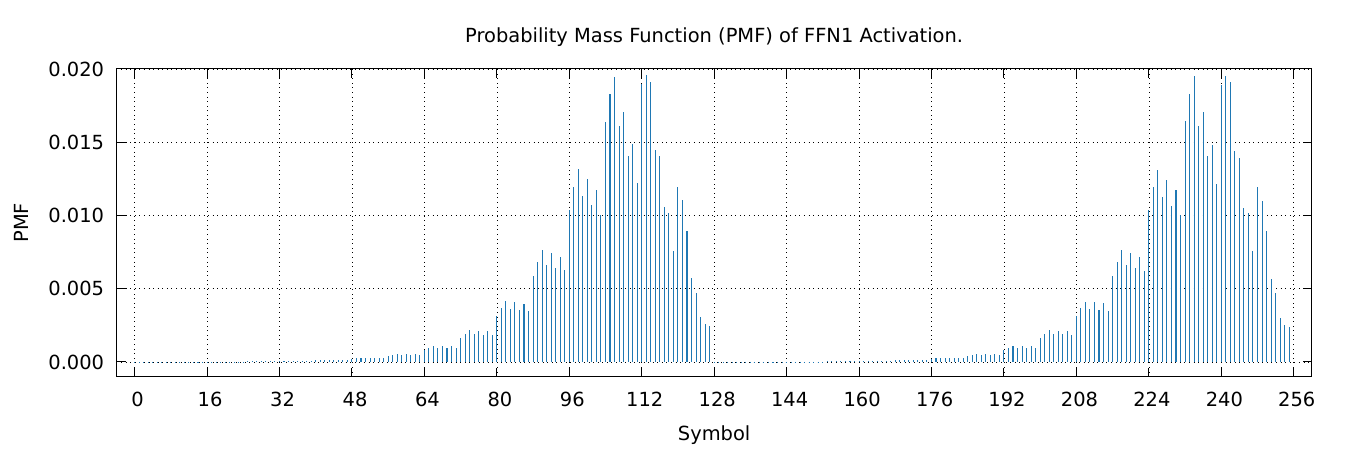}
  \caption{Probability Mass Function (PMF) of FFN1 activation.}
  \label{fig:ffn1_pmf}
\end{figure}

Fig. \ref{fig:ffn1_sorted_pmf} and 
Fig. \ref{fig:ffn2_sorted_pmf}
show the sorted PMF of FFN1 and FFN2 activations.
Fig. \ref{fig:ffn1_pmf} shows the actual
probability distribution of FFN1 activation.
The symbols with the highest frequencies are
113, 241, 234, 106 etc.
The symbols with the lowest frequencies are
141, 137, 0, 128 etc.

In a software or hardware implementation,
the encoder maintains a 256 entry Look Up Table
(LUT) as shown in Table \ref{table:ffn1_encoder_lut}.
While creating the table, the input symbols are sorted
in the order of decreasing probabilities. They are mapped
to symbols 0 through 255 and assigned a code according
to the coding scheme of
Table \ref{table:ffn1_quad_length} or
Table \ref{table:ffn2_quad_length}.
After the table is populated, the input symbols
i.e., the first column is sorted in ordinal
fashion to facilitate indexing into the LUT.
The second column is shown only for convenience.
During encoding, we simply lookup the code
corresponding to the input symbol.

\begin{table}[h]
\caption{Encoder Look Up Table.}
\renewcommand{\arraystretch}{1.5}
\centering
\rowcolors{1}{RoyalBlue!10}{}
\begin{tabular}{|c|c|l|}
\toprule \hline
\rowcolor{RoyalBlue!40}
Input Symbol    & Mapped to Symbol  & Code     \\
113             & 0                 & 000\_000 \\
241             & 1                 & 000\_001 \\
234             & 2                 & 000\_010 \\
\dots           & \dots             & \dots      \\
233             & 8                 & 001\_000 \\
\dots           & \dots             & \dots      \\
137             & 253               & 111\_11111101 \\
0               & 254               & 111\_11111110 \\
128             & 255               & 111\_11111111 \\
\hline \bottomrule
\end{tabular}
\label{table:ffn1_encoder_lut}
\end{table}

During decoding, we use the `area code' i.e.,
the first three bits to determine the length of the code.
Using the coding scheme from Table \ref{table:ffn1_quad_length},
if the area code is $000$ through $100$, we read the next 3 bits.
If the area code is $101$, we read the next 4 bits and so on.
Then we add the appropriate offset to obtain the encoded symbol.
For example, if the area code is $100$ and the next 3 bits are
$010$, then the encoded symbol is $32 + 2 = 34$.
The decoder maintains a 256 entry LUT to convert
the encoded symbol to the output symbol as shown in
Table \ref{table:ffn1_decoder_lut}.
Note that multiple LUTs, one for each tensor type e.g., FFN1
activation, FFN1 activation gradient etc., can be
obtained apriori as shown in \cite{single_stage_huffman}.

\begin{table}[h]
\caption{Decoder Look Up Table.}
\renewcommand{\arraystretch}{1.5}
\centering
\rowcolors{1}{RoyalBlue!10}{}
\begin{tabular}{|c|c|l|}
\toprule \hline
\rowcolor{RoyalBlue!40}
Encoded Symbol  & Output Symbol \\
0               & 113   \\
1               & 241   \\
2               & 234   \\
\dots           & \dots \\
8               & 233   \\
\dots           & \dots \\
253             & 137   \\
254             & 0     \\
255             & 128   \\
\hline \bottomrule
\end{tabular}
\label{table:ffn1_decoder_lut}
\end{table}


\section{Conclusion \& Future Work}

Lossless compression is an effective way to reduce network
traffic and improve collective performance.
Huffman codes are optimal entropy codes but suffer from
slow bit-sequential decoding.
This paper introduced Quad Length Codes, a coding
scheme for lossless compression of e4m3 data
type in ML applications.
The scheme exploits the symbol frequency distributions
similar to Huffman codes but trades off some of its
compression efficiency for decoding speed and
implementation simplicity.

The coding scheme uses 3 prefix bits to divide 
the 256 symbols into 8 areas. Each area has a
different number of symbols and hence has a different
code length. The scheme uses a Look Up Table with
256 entries, significantly reducing the
implementation complexity compared to Huffman trees.
While a compressibility of 13.9\% achieved using
these codes is slightly lower than the 15.9\%
achieved by Huffman codes, Quad Length Codes
reduce the latency and complexity associated
with Huffman decoding.

Our coding schemes were obtained empirically.
It is possible to tweak the number of areas,
the number of symbols in each area, and
the number of unique code lengths
to achieve a better compression ratio
for the above and other distributions.
In future, we want to develop a 
mathematical formulation of the problem.

\bibliography{citation}

@misc{deflate,
    title={{DEFLATE}},
    url={https://en.wikipedia.org/wiki/Deflate},
}

@misc{huffman,
    title={{Huffman Coding}},
    url={https://en.wikipedia.org/wiki/Huffman_coding},
}

@misc{zstd,
    series =    {Request for Comments},
    number =    8878,
    howpublished =  {RFC 8878},
    publisher = {RFC Editor},
    doi =       {10.17487/RFC8878},
    url =       {https://www.rfc-editor.org/info/rfc8878},
    author =    {Yann Collet and Murray Kucherawy},
    title =     {{Zstandard Compression and the 'application/zstd' Media Type}},
    year =      2021,
    month =     Feb,
}

@article{brotli,
author = {Alakuijala, Jyrki and Farruggia, Andrea and Ferragina, Paolo and Kliuchnikov, Eugene and Obryk, Robert and Szabadka, Zoltan and Vandevenne, Lode},
title = {{Brotli: A General-Purpose Data Compressor}},
year = {2018},
publisher = {Association for Computing Machinery},
address = {New York, NY, USA},
url = {https://doi.org/10.1145/3231935},
journal = {ACM Trans. Inf. Syst.},
month = Dec,
}

@misc{universal,
    title={{Universal Code}},
    url={https://en.wikipedia.org/wiki/Universal_code_(data_compression)},
}

@misc{gamma-elias,
    title={{Elias Gamma Coding}},
    url={https://en.wikipedia.org/wiki/Elias_gamma_coding},
}

@misc{delta-elias,
    title={{Elias Delta Coding}},
    url={https://en.wikipedia.org/wiki/Elias_delta_coding},
}

@misc{omega-elias,
    title={{Elias Omega Coding}},
    url={https://en.wikipedia.org/wiki/Elias_omega_coding},
}

@misc{exp-golomb,
    title={{Exponential-Golomb Coding}},
    url={https://en.wikipedia.org/wiki/Exponential-Golomb_coding},
}

@misc{single_stage_huffman,
    title={{Single-Stage Huffman Encoder for ML Compression}},
    author={Aditya Agrawal and Albert Magyar and Hiteshwar Eswaraiah and Patrick Sheridan and Pradeep Janedula and Ravi Krishnan Venkatesan and Krishna Nair and Ravi Iyer},
    year={2026},
    url={https://arxiv.org/abs/2601.10673},
}

@misc{entropy,
    title={{Entropy}},
    url={https://en.wikipedia.org/wiki/Entropy_(information_theory)},
}

@misc{format:exmy,
    title={{eXmY: A Data Type and Technique for Arbitrary Bit Precision Quantization}}, 
    author={Aditya Agrawal and Matthew Hedlund and Blake Hechtman},
    year={2024},
    url={https://arxiv.org/abs/2405.13938},
}

@misc{format:ocpmx,
    title = {{OCP Microscaling Formats (MX) Specification}},
    url = {https://www.opencompute.org/documents/ocp-microscaling-formats-mx-v1-0-spec-final-pdf},
    author = {Bita Darvish Rouhani and Nitin Garegrat and Tom Savell and Ankit More and others},
    altauthor = {Bita Darvish Rouhani and Nitin Garegrat and Tom Savell and Ankit More and Kyung-Nam Han and Colin Verrilli},
    year = {2023},
    month = {Sep},
}

@misc{models:gemma1,
    title={{Gemma: Open Models Based on Gemini Research and Technology}},
    author={Thomas Mesnard and Cassidy Hardin and Robert Dadashi and others},
    altauthor={Gemma Team and Thomas Mesnard and Cassidy Hardin and Robert Dadashi and Evan Senter and Alek Andreev and Kathleen Kenealy},
    year={2024},
    url={https://arxiv.org/abs/2403.08295},
}

@misc{models:gemma2,
    title={{Gemma 2: Improving Open Language Models at a Practical Size}},
    author={Morgane Riviere and Shreya Pathak and Pier Giuseppe Sessa and others},
    altauthor={Gemma Team and Morgane Riviere and Shreya Pathak and Pier Giuseppe Sessa and Cassidy Hardin and Surya Bhupatiraju and Léonard Hussenot and Alek Andreev},
    year={2024},
    url={https://arxiv.org/abs/2408.00118},
}

@misc{models:gemini_2025,
    title={{Gemini: A Family of Highly Capable Multimodal Models}},
    author={Rohan Anil and Sebastian Borgeaud and Jean-Baptiste Alayrac and Jiahui Yu and Radu Soricut and others},
    year={2025},
    url={https://arxiv.org/abs/2312.11805},
}

@misc{models:llama,
    title={{LLaMA: Open and Efficient Foundation Language Models}}, 
    author={Hugo Touvron and Thibaut Lavril and Gautier Izacard and Xavier Martinet and Marie-Anne Lachaux and Timothée Lacroix and Baptiste Rozière and Naman Goyal and Eric Hambro and Faisal Azhar and Aurelien Rodriguez and Armand Joulin and Edouard Grave and Guillaume Lample},
    year={2023},
    url={https://arxiv.org/abs/2302.13971}, 
}

@misc{models:gpt_1,
    title={{Improving Language Understanding by Generative Pre-Training}}, 
    author={Alec Radford and Karthik Narasimhan and Tim Salimans and Ilya Sutskever},
    year={2018},
    url={https://cdn.openai.com/research-covers/language-unsupervised/language_understanding_paper.pdf},
}

@misc{parallelism:nvidia_1,
    title={{Parallelisms Guide}},
    url={https://docs.nvidia.com/nemo/megatron-bridge/0.2.0/parallelisms.html},
}

@misc{parallelism:nvidia_2,
    title={{Mastering LLM Techniques: Inference Optimization}},
    author={Shashank Verma and Neal Vaidya},
    year={2023},
    url={https://developer.nvidia.com/blog/mastering-llm-techniques-inference-optimization/},
}

@misc{parallelism:colossalai,
    title={{Paradigms of Parallelism}},
    author={Shenggui Li and Siqi Mai},
    url={https://colossalai.org/docs/concepts/paradigms_of_parallelism/},
}

@article{parallelism:scaling-book,
      title = {{How to Scale Your Model}},
      author = {Austin, Jacob and Douglas, Sholto and Frostig, Roy and Levskaya, Anselm and Chen, Charlie and Vikram, Sharad and Lebron, Federico and Choy, Peter and Ramasesh, Vinay and Webson, Albert and Pope, Reiner},
      url = {https://jax-ml.github.io/scaling-book/},
      year = {2025},
}

@misc{parallelism:zero,
    title={{ZeRO: Memory Optimizations Toward Training Trillion Parameter Models}}, 
    author={Samyam Rajbhandari and Jeff Rasley and Olatunji Ruwase and Yuxiong He},
    year={2020},
    url={https://arxiv.org/abs/1910.02054},
}

@misc{parallelism:megatron,
    title={{Megatron-LM: Training Multi-Billion Parameter Language Models Using Model Parallelism}},
    author={Mohammad Shoeybi and Mostofa Patwary and Raul Puri and Patrick LeGresley and Jared Casper and Bryan Catanzaro},
    year={2020},
    url={https://arxiv.org/abs/1909.08053},
}

@misc{parallelism:sequence,
    title={{Reducing Activation Recomputation in Large Transformer Models}}, 
    author={Vijay Korthikanti and Jared Casper and Sangkug Lym and Lawrence McAfee and Michael Andersch and Mohammad Shoeybi and Bryan Catanzaro},
    year={2022},
    url={https://arxiv.org/abs/2205.05198},
}

@misc{collectives:nccl,
    title={{Collective Operations}},
    url={https://docs.nvidia.com/deeplearning/nccl/user-guide/docs/usage/collectives.html},
}

\end{document}